\documentclass[conference]{IEEEtran}
\IEEEoverridecommandlockouts
\usepackage{cite}
\usepackage{amsmath,amssymb,amsfonts}
\usepackage{algorithmic}
\usepackage{graphicx}
\usepackage{textcomp}
\usepackage{xcolor}
\def\BibTeX{{\rm B\kern-.05em{\sc i\kern-.025em b}\kern-.08em
    T\kern-.1667em\lower.7ex\hbox{E}\kern-.125emX}}

\usepackage{algorithm2e}
\usepackage{ dsfont }
\usepackage{makecell}
\usepackage{arydshln}
\usepackage{hyperref}

\begin{document}

\title{XAI-N: Sensor-based Robot Navigation using Expert Policies and Decision Trees}

\author{\IEEEauthorblockN{Aaron M. Roth}
\IEEEauthorblockA{\textit{Department of Computer Science} \\
\textit{University of Maryland}\\
College Park, Maryland, USA \\
amroth@umd.edu}\thanks{This work was supported in part by ARO Grants W911NF1910069, W911NF2110026  and U.S. Army Grant No. W911NF2120076}
\and
\IEEEauthorblockN{Jing Liang}
\IEEEauthorblockA{\textit{Department of Computer Science} \\
\textit{University of Maryland}\\
College Park, Maryland, USA }
\and
\IEEEauthorblockN{Dinesh Manocha}
\IEEEauthorblockA{\IEEEauthorblockA{\textit{Department of Computer Science} \\
\textit{University of Maryland}\\
College Park, Maryland, USA }} }

\maketitle

\begin{abstract}
We present a novel sensor-based learning navigation algorithm to compute a collision-free trajectory for a robot in dense and dynamic environments with moving obstacles or targets. Our approach uses deep reinforcement learning-based expert policy that is trained using a sim2real paradigm. In order to increase the reliability and handle the failure cases of the expert policy, we combine with a policy extraction technique to transform the resulting policy into a decision tree format. 
%The resulting decision tree has properties which 
We use properties of  decision trees to analyze and modify the policy and improve performance of navigation algorithm including smoothness, frequency of oscillation, frequency of immobilization, and obstruction of target. Overall, we are able to modify the policy to design an improved learning algorithm without retraining.  We highlight the benefits of our approach in simulated environments and navigating a Clearpath Jackal robot among moving pedestrians. (Videos at this url: \url{https://gamma.umd.edu/researchdirections/xrl/navviper}) 
\end{abstract}

%\begin{IEEEkeywords}
%Decision Trees, Motion and Path Planning, Motion Control, Reactive and Sensor-Based Planning 
%\end{IEEEkeywords}

Learning methods are increasingly being used for robot navigation. Methods including Deep Reinforcement Learning (DRL)~\cite{arulkumaran2017deep}, learning from demonstration~\cite{argall2009survey}, imitation learning, etc. are able to integrate well with sensor data and have been used for 
navigation in % DEL-OPT
real-world scenarios.
They can work well in %many challenging scenarios corresponding to
dense environments with multiple dynamic obstacles.
However, %a number of issues remain. Training a DRL policy can take large number of simulated timesteps~\cite{fan2018crowdmove}. %PHD   RE-ADD
when trying out a policy in a new environment or with a different configuration of obstacles, it can fail in simulation or in the real world,  %TODO-REF
with failure modes including collisions,  %TODO-REF
oscillatory behaviors/non-smooth paths,  %TODO-REF
or 
agent-induced immobilization (``freezing''),  %TODO-REF
among others~\cite{kahan2009backup, brooks2017human, jain2018evaluating, lee2010gracefully, morales2019interaction, kontogiorgos2020embodiment, ramanagopal2018failing}. % TODO Distribute
% CAMERA- add in Towards Robust...
%  and the best methods can still encounter errors.

The policies that result from a learning method like DRL are typically opaque as to their inner workings, and cannot easily be directly analyzed or modified without revising the method and repeating the time-consuming training step.  
% This ``black box'' nature means that when the trained policies fail, and when error cases are discovered, there is often no recourse but to retrain from scratch.  %PHD
Furthermore, it is hard to predict when errors would occur without running the policy.%, or have any guarantees about when it will or won't perform as expected.  %PHD
This makes it difficult to have any confidence or reliability in future performance, especially if the operating environment differs from the training data.
%
%As a result, it is important to identify and correct the aforementioned behavioral issues without retraining and without requiring an optimal learning procedure.

%As a motivating example, consider an engineer using a policy learned over the course of a month via a complex procedure to which they may not have access. This person should be able to adjust that policy to suit their custom needs.  Most existing methods primarily focus on the initial learning stage, and not on what comes afterwards.
% As a motivating example, consider an engineer end-user faced with a policy learned over the course of a month via a complex procedure to which they may not have access and of the workings of which they may not have understanding. This person should not be prohibited from adjusting that policy to suit their custom needs.  However, that is precisely the situation that could arise as a result of most methods, which are primarily focused on the initial learning stage, and not on what comes afterwards. %PHD

There is considerable interest in developing explainability and 
interpretability in deep learning and reinforcement learning methods~\cite{doshi2017towards,sado2020explainable}. The ultimate goal is to develop methods and AI techniques such that the results of the solution can be understood by humans. 
This is in contrast with the most widely used % DEL-OPT
machine learning methods that tend to act like a black box and even the designers cannot explain why the underlying method arrived at a specific decision. Our main goal is to design explainable algorithms for robot navigation, where we can offer some insights about their performance in different scenarios. In this context, we address the problem of modifying the policy to improve the  performance of learning-based navigation methods.

%While there are good theoretical developments, 

%In this paper, we focus on using techniques from the broad field of Explainable AI to 

%z5learning methods for different applications
%As our technique helps a technical end-user modify and work with a learned policy, it is related to the %  TODO1-(some other area of research like practical robotics or robo for industry)
% To address similar concerns in the broader machine learning community, there is a
%developing body of work focusing on %explainability and 
%interpretability in deep learning and reinforcement learning~\cite{doshi2017towards,sado2020explainable}. We %go a step beyond simply gaining insight, and
%ask how can we can enable a team to improve a policy.
%There are a variety of approaches.  Some involve transforming a non-interpretable policy into an interpretable structure~\cite{ross2011reduction,VIPER}.  Others include using an interpretable component alongside a non-interpretable component, such as using a tree for regularization~\cite{wu2018beyond}.
% One of our main goals is to develop good learning-based navigation algorithms based on recent advances in explainability and interpretability. 

\begin{figure}
    \centering
    \centering
    {{\includegraphics[width=4cm,height=4cm]{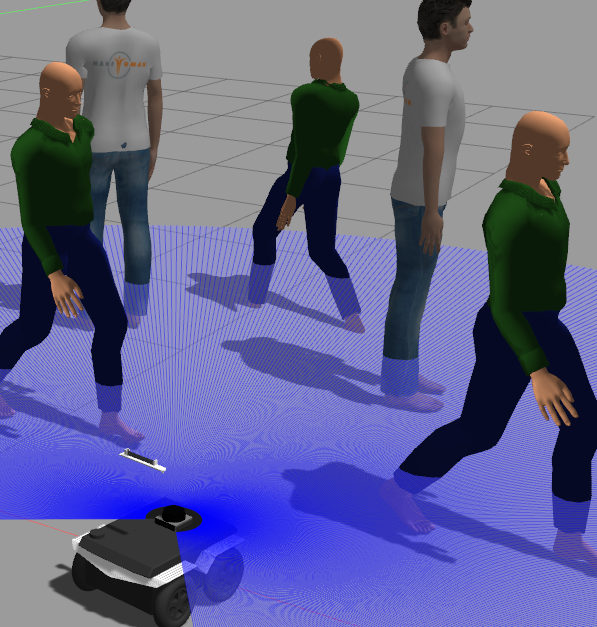} }}%
    % \qquad
    % \subfloat \centering {{\includegraphics[width=4cm,height=4.5cm]{images/narrowpassage.png} }}
    {{%\includegraphics[width=4cm,height=4cm]{images/realworld_jackal2.png} 
    \includegraphics[width=4cm,height=4cm]{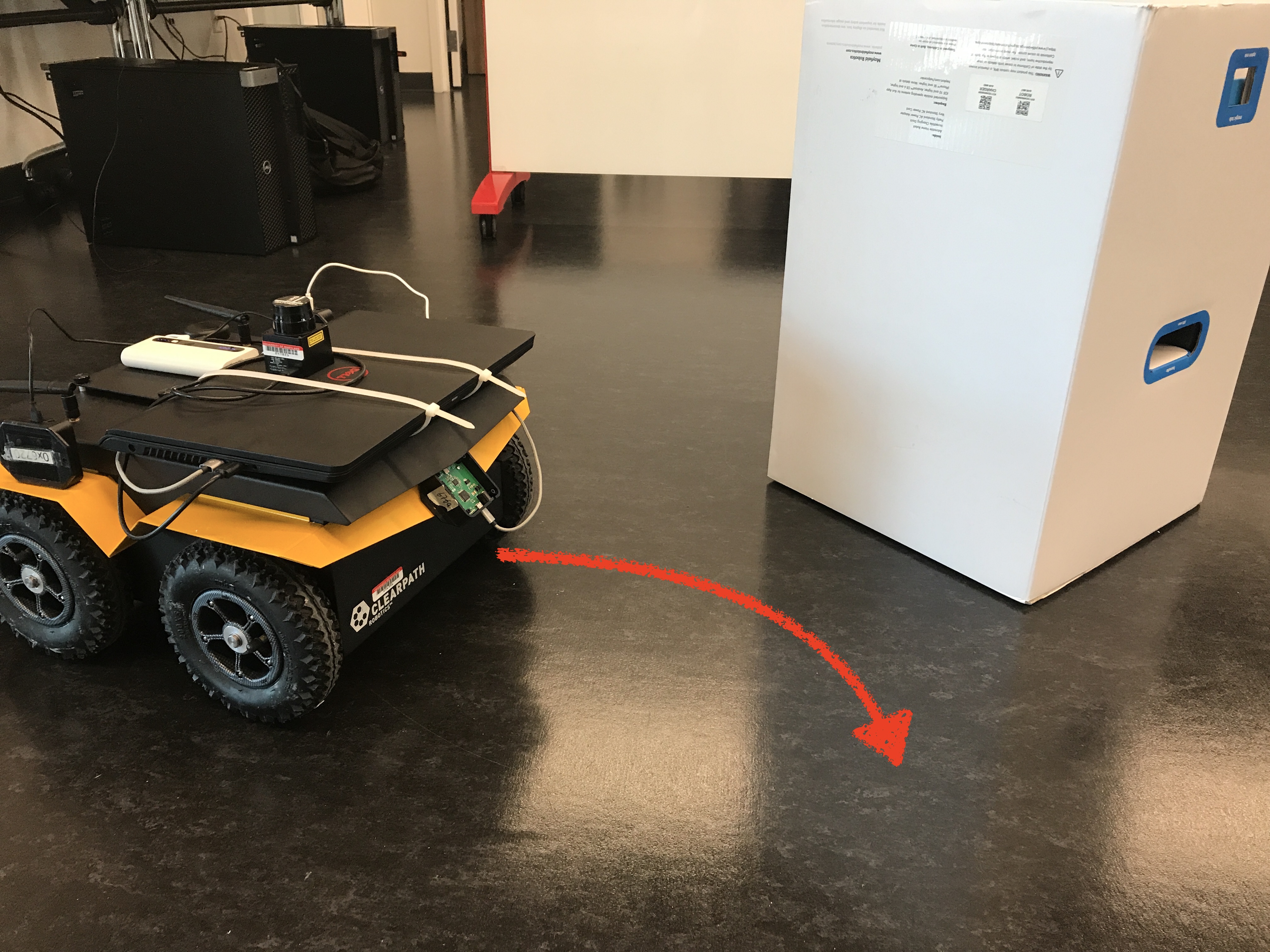}
    }}
    \caption{Robot Navigation using XAI-N: The left image corresponds to training scenarios used for training a DRL policy; right image is testing of the 
    % interpretable 
    tree policy demonstrated on a Clearpath Jackal robot. XAI-N generates an interpretable tree policy that allows us to identify and fix failure modes of the DRL policy without retraining. This results in improved navigation behavior, including fewer oscillations or freezing problems in dynamic environments.}
    \label{fig:frontpage}
\end{figure}

\noindent {\bf Main Results:} We introduce a novel scheme, XAI-N, which integrates the concepts of expert policies, policy extraction, and decision trees and utilizes them to make policy modifications that improve navigation in dynamic environments.
% we propose a scheme, convert into decision tree with policy extraction; policy modification
% these two stages result in better understanding of tree, can make assertions about the tree (how much certain type of freezing).  better explainability and interpetability.  explainability and interpetability leads us to learn better navigation methods
% We address this issue with our hybrid learning approach.
Our approach first learns a navigation policy using DRL. 
We then transform this neural net policy into a decision tree (DT) policy using an imitation learning policy extraction method. A decision tree is a flowchart like tree structure (a binary tree in our case) that classifies (or maps) a space of numerical features into subsections (leaves) corresponding to classes. We use a tree where the features correspond to sensor inputs and other aspects of a state space and the leaf classifications correspond to discrete action choices, allowing us to use a tree as policy for robot navigation. DTs are inherently interpretable, as every output of a decision tree is tractable~\cite{hall1998decision}. Once transformed into a DT, we show the navigation policy's structure can be analyzed, interpreted, and modified to design an improved navigation algorithm.

We take advantage of % interpretable properties of 
the tree structure to detect and address suboptimalities in the policy and improve the navigation across several metrics: i) smoothness/oscillation frequency, %ii) time-to-goal, 
ii) freezing frequency, iii) total path length, iv) blocking/obstruction occurrence, and v) reward per timestep (a scalar measure combining multiple of the preceding metrics). In this manner, we get the best of both worlds--the ability of the DRL to learn complex tasks and handle sensor data, and the %interpretability and 
comprehensible malleability of the decision tree.  
%Our work combines hand-crafted algorithmic approaches with autonomous reinforcement learning (RL).
The  novel contributions of our paper include
\begin{enumerate}
    \item XAI-N, robot navigation learning method that combines traditional DRL learning with rule-based domain-specific algorithms. %by using policy extraction to decision trees%, and analysis and moifcation of the latter %, resulting in an interpretable navigation policy.
    % \item Improve explainability and interpretability of black-box learned policies in Robot Navigation
    \item Take advantage of the extracted tree structure to improve overall navigation scheme:
    % explainable aspects of approach to improve overall navigation scheme:
    \begin{enumerate}
        \item %Without executing the policy, 
        Detect situations that could cause ``freezing'' 
        %(when the robot is rendered immobile, unable to find the valid existing path) 
        and modify policy to preclude such failure cases
        \item Observe when oscillation occurs and modify  policy to smooth the path, decreasing oscillation.
        \item Prevent robot obstructing a human it is following
    \end{enumerate}
  %  \item Demonstrate the benefits in real and synthetic scenarios, where a robot has access only to information from its sensors.   (a) and (b) tested on a small robot navigating to a goal location amoung dense crowds, and (c) on a game-like agent finding and following a human in a warehouse
    % \item  an open source code for running navigation tasks with the jackal formulated as an OpenAI environment (and thus compatible with many standard RL tools)
\end{enumerate}
We highlight the benefits of our approach in many simulated scenarios and on a Clearpath Jackal robot navigating among obstacles and pedestrians.  %We also compare its performance with other sensor-based robot navigation algorithms and demonstrate improvements in terms of different metrics.

\section{Related Work}\label{sec:rw} % 1.5 column

\subsection{Learning for Navigation}

The last decade has seen the rise of learning-based robot navigation algorithms~\cite{WB1,JHow2,JiaPan1,sathyamoorthy2020densecavoid}, which can directly handle the real-world representations captured using commodity visual sensors.
This enhances the ability of a robot to adapt and reach the goal even in new, unknown environments.  Some of the widely used methods are based on  Deep Learning (DL) or Reinforcement Learning (RL)~\cite{mirowski2016learning,zhang2017deep,JHow2}.
% However, in case of DL based methods, it is difficult to collect a large set of labeled data for training deep neural networks for navigation in unknown environments, and hence this leads to poor generalization. On the other hand, RL methods involve learning using the experiences gained from the interaction of the robot with its environment and  can be generalized. Zhu et al.\cite{zhu2017target, kulhanek2019vision} proposed The House Of inteRactions (AI2-THOR) platform and used camera to detect a robot's goal and current positions, and trained a policy by actor-critic protocol for navigation. 
%  can remove below for space reasons
Xie et al.\cite{xie2017towards} trained a network to convert RGB images to depth images and then used deep double-Q network(D3QN) algorithm to navigate the robot avoiding collisions~\cite{govindaraju2005quick}. %Although those methods have good performance in avoiding static obstacles, it is challenging to deal with dynamic scenes.
In order to perform dynamic obstacles avoidance, Everett et al.\cite{JHow2} proposed a strategy, GPU/CPU Asynchronous Advantage Actor-Critic for collision avoidance with Deep RL(GA3C-CADRL), using LSTM. L{\"o}tjens et al.~\cite{lotjens2019safe} developed an uncertainty-aware navigation method to  avoid pedestrians. 
A common limitation of all these learning methods is that the black-box properties of neural networks make it hard to modify them, except by attempting to retrain or develop an improved learning method.

% A common theme across all these learning methods is that despite showing promising results on different navigation tasks, the black-box properties of neural networks makes it hard to evaluate them on the basis of safety and reliability.  The interpretation of learning method is necessary not only for understanding decision making process but also for finding reasonable methods to alleviate existing or unknown issues. Our NAV-VIPER addresses some of these issues.   

% \subsection{XAI Learning Methods}

% Motivated by the desire to understand the sometimes opaque and inscrutabble nature of many advanced deep learning methods, Explainble AI (XAI) is a growing area of exploration~\cite{doshi2017towards, sado2020explainable}. 

% \subsection{Combined Learning Methods}

% The concept of ``hybrid learning'' is the idea of combining multiple approaches in parallel or in sequence to achieve some learning goal.~\cite{abraham2009hybrid}
% There are previous techniques that combine reinforcement learning and imitation learning.~\cite{gao2018reinforcement, ziebart2008maximum, le2018hierarchical}
% In contrast to the most existing approaches, which seek to create more powerful fully-autonomous learning agents, we are interested in combining learning approaches with rule-based, algorithmic procedures. This reflects the real world needs of robot developers.

% We combine reinforcement learning (mentioned above) with policy extraction via imitation learning (described below) and decision tree modification (which we introduce in this work) to produce XAI-N.

\subsection{Policy Extraction and Imitation Learning}

Initiation Learning ~\cite{hussein2017imitation, ho2016generative} involves learning a policy via copying an existing ``expert'' policy or deriving a policy that best fits observed procedure (learning from demonstration ~\cite{argall2009survey}). 
Policy Extraction (also called Policy Distillation) is the process of taking an existing trained policy and transforming it into a different format. This could be transforming a neural network into a smaller neural network ~\cite{rusu2015policy} or turning a neural network into some other format such as a tree.~\cite{jhunjhunwala2019policy}
% In our work, we use imitation learning to accomplish policy extraction, turning a neural net policy into a decision tree policy.
VIPER~\cite{VIPER} is an algorithm which learns an ``expert'' policy using a neural net (such as PPO ~\cite{PPO}) and then uses imitation learning to fit a decision tree to replicate the expert policy. VIPER has been used to generate decision tree policies for proof-of-concept problems such as CartPole~\cite{bhupatiraju2018towards}, Atari Pong~\cite{VIPER} and other simulations such as CARLA~\cite{chen2020learning}. Our approach also uses VIPER.

% Most previous works involving policy extraction stop there--we utilize the structure of the extracted policy to perform additional improvements. %PHD

\subsection{XAI and Analyzing or Utilizing Decision Trees}
Motivated by the desire to understand the sometimes opaque and inscrutabble nature of many advanced deep learning methods, Explainble AI (XAI) is a growing area of exploration~\cite{doshi2017towards,sado2020explainable}. 
One class of XAI methods is that of globally intrinsic~\cite{alharin2020reinforcement} explanation methods, such as decision trees.
There is prior work on using a directly interpretable structure such as a tree or graph~\cite{Roth_MS_Thesis2019}. Previous authors have used decision trees in conjunction with RL.  
A deep neural network can be distilled into a soft decision tree~\cite{frosst2017distilling}, or learned via RL using Policy Tree~\cite{das2015adaptive}. However, neither of these methods are interpretable. Some methods such as the Pyeatt Method ~\cite{pyeatt2003reinforcement} and Conservative Q-Improvement~\cite{roth2019conservative} use an RL method to learn a decision tree in an additive manner.
Decision trees, while hard to learn, are attractive as policies because they yield benefits in terms of interpretation and verifiability.
DT are also well-suited for safety-critical applications because there are a range of standard techniques (such as Z3~\cite{de2008z3}) that can be used to perform verification analysis on them~\cite{blanchet2003static}.

% Theoretically, trees could be modified based on domain-specific algorithms, since the relationship between features and actions is clear and easily modifiable.  We pursue this avenue of investigation in this work. To our knowledge, using domain knowledge to modify a tree that was originally constructed using a general learning objective (as in the reward and loss in DRL) has not been investigated in the context of Robot Navigation.

There is work on modifying decision trees to better fit a dataset, such a simplification~\cite{breslow1997simplifying} or pruning ~\cite{eggermont2004detecting}.  This could result in loss of accuracy, and it is more about changing structure without impeding performance than it is about improving performance. There is also work on adapting a DT from one task to another.~\cite{won2007transfer}
Excluding a paper on classification~\cite{aitkenhead2008co}, and retraining a tree after modifying a dataset%TODO-REF-NEW
, we found no prior work on tree modification for the purpose of addressing a specific domain goal as we do. % (ie targetting freezing or oscillation or other errors). %(The exception is if you count things that involve retraining the tree--ie modifying dataset and retraining, or otherwise making a decision to retrain in a 'smarter' way. There seems to be much work of that nature.)
% are we the first to?  %PHD
% nothing about improving performance using domain knowledge outside original dataset or optimziing for a secondary metric  

\section{Problem Setup and Overview}\label{sec:prob} % (0.5 column)
% Notation table
\subsection{Problem Setup}
We model our navigation task as a Partially Observable Markov Decision Process (POMDP), which is represented by a tuple $(Z, S, A, P, R, \gamma)$. $Z$ is the real state space, $S$ is the observed state space, $A$ is the action space, $R$ is rewards, and $P$ is the state transition dynamics: $S \times A \xrightarrow{} S$, and $\gamma \in (0,1 ] $ is a discount factor. % The distribution of initial state is represented as $\rho_0$. 
Our goal is to generate an optimal policy $\pi$ which maximizes the discounted reward function:
\begin{equation}
        \eta(\pi) = \mathbb{E}_{\pi}\left[\sum^{T-1}_{t=0}{\gamma^{t} r(s_t,a_t)}\right].
\end{equation}
% Our goal is to develop a methodology to facilitate combining the advantages of interpretable and black-box methods, and we explore this goal in the context of a robotics crowd navigation task.  PHD
The goal of the task is to make a robot learn to efficiently go to its goal position and at the same time avoid collisions with obstacles. 
Obstacles can be static or dynamic. %with relatively slow speed, which can be captured by a robot's sensors.
The robot performs local navigation using only what is observed by the sensors and knowledge of the most recently taken action. %For observation, we assume the robot can only observe obstacles by using 2D lidar which has a range of detection within a few meters.
At each step, a robot knows the goal position relative to itself (due to sensors). The episode ends when the goal state is reached (to within a tolerance) or a collision occurs. We develop an XAI method to address this type of robot navigation problem.

\begin{figure}
    \centering
    \includegraphics[width=0.4\textwidth]{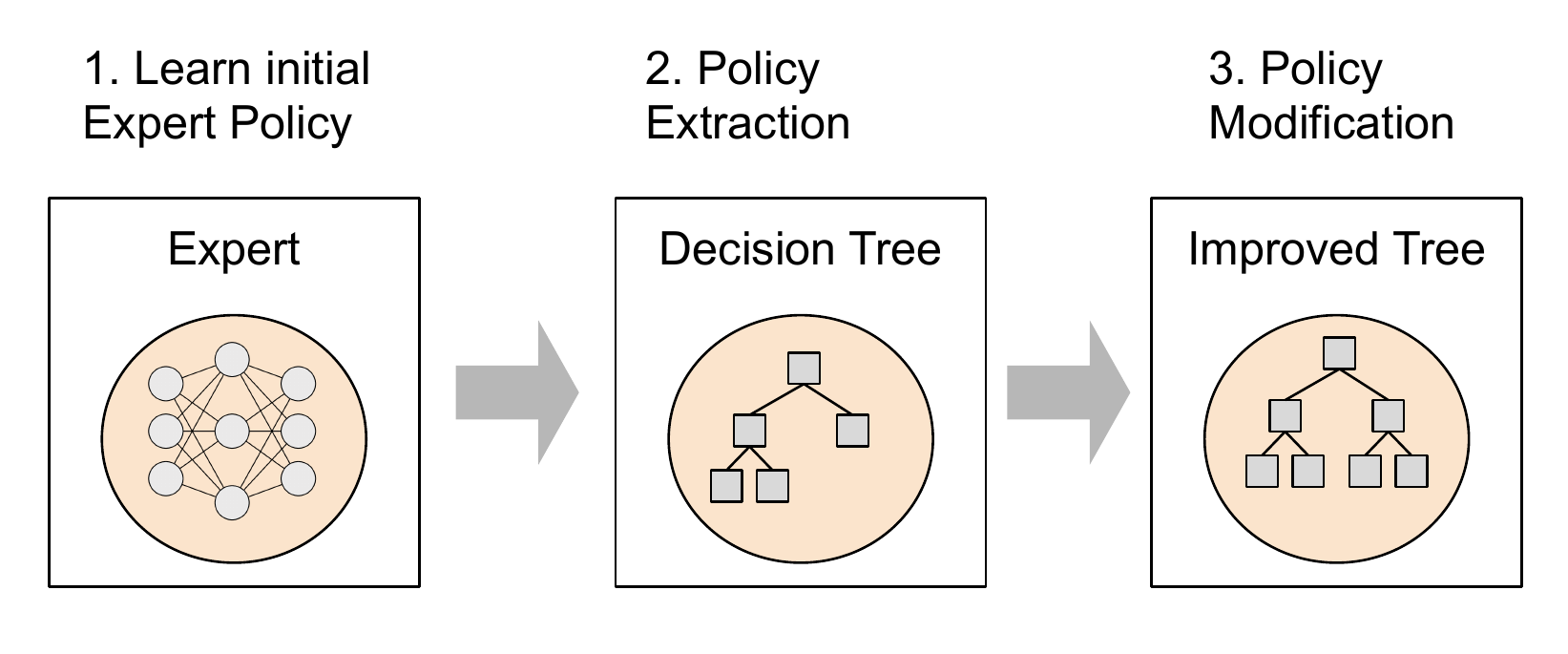}
    \caption{Our XAI-N Robot Navigation Algorithm with three stages: In stage 1, we train an expert policy $\pi^*$ with a learning method such as DRL. In stage 2, we perform policy extraction to decision tree policy $\pi^\dag$. In stage 3, we apply modifications (such as the oscillation fix and freezing fix) to correct errors and improve the policy with regards to navigation metrics. This results in an improved interpretable navigation policies as compared to DRL without retraining.}
    \label{fig:paradigm}
\end{figure}

\begin{table}[b]
    \centering
    \resizebox{0.5\textwidth}{!}{
    \begin{tabular}{|c|l|}
\hline
$s \in S$ & \makecell[l]{a state $s$ is an array representing the state of the world and the robot in it} \\
\hline
$a \in A$ & \makecell[l]{an action $a$ is a single discrete action in the set of possible actions $A$} \\
\hline
$a_F$ & the ``stop'' action\\
\hline
$a_L$ & a ``rotate left'' action\\
\hline
$a_R$ & a ``rotate right'' action\\
\hline
$a_D$ & a ``forward'' action\\
\hline
$\mathbf{C}_F$ & a set of column indices of polar columns in front of robot\\
\hline
$\mathbf{A}_D$ & the set of actions containing a component of forward movement\\
\hline
$P$ & state transition dynamics\\
\hline
$r, R$ & reward (for a single ($s$, $a$) pair or in general, respectively)\\
\hline
$\gamma$ & future discount factor\\
\hline
$E$ & \makecell[l]{An environment (real or simulated). Receives an action $a$ and provides the\\perceived state of the world $s$, reward $r$, and boolean indication of \\whether the goal has been reached (``done'').}\\
\hline
$\pi: (s \rightarrow a) $ & a policy, mapping a state $s$ to action $a$\\
\hline
$\pi^*$ & an expert policy (neural net in our work, but can be anything)\\
\hline
$\pi^\dag$ & a decision tree policy\\
\hline
$F^\dag(\pi)$ & policy extraction conversion function, outputs $\pi^\dag$\\
\hline
$m_A$ & \makecell[l]{how much movement to allow in a ``static'' $s$ during freezing detection}\\
\hline
$\mathcal{F}_O$ & \makecell[l]{takes a history of $(s, a)$ pairs and outputs boolean indicating whether\\ oscillation has occurred or not}\\
\hline
$\mathcal{N}$ & a set of nodes with errors detected\\
\hline
\makecell{$\mathcal{O}_C(i)$,\\where $n_i \in \mathcal{N}$} & set of states in state subspace of node $i$ where oscillation occurs\\
\hline
\makecell{$\mathcal{O}_X(i)$\\where $n_i \in \mathcal{N}$}& set of states in state subspace of node $i$ where oscillation does not occur\\
\hline
\end{tabular}
}
    \caption{Symbols and Notation used in the paper}
    \label{tbl:notation}
\end{table}
%old from other paper The subsequent goals are to (i) learn a decision tree representation of this policy (such that the mapping from state to action is in the form of a tree instead of a network (see Section \ref{sec:approach-DT} for details)) with minimal loss of performance, and to then (ii) take advantage of the tree structure to improve the policy using methods that could not as easily be applied to a pure neural net.

\subsection{XAI Robot Navigation Algorithm Overview}\label{sec:app-paradigm}

A diagram of our XAI-N process is shown in Figure \ref{fig:paradigm}. In the first stage, a robot navigation policy is learned as an ``expert policy''.
% according to the user's choice of the existing best methods.  
This process is discussed briefly in Section \ref{sec:app-box1}, and the resulting policy is referred to as the ``expert policy.''  Next, some appropriately chosen Policy Extraction~\cite{rusu2015policy, jhunjhunwala2019policy}
or Imitation Learning~\cite{hussein2017imitation} 
process is used to transform the expert policy into a decision tree format policy. A DT policy 
$\pi^\dag$ uses a DT to perform the mapping of state $s$ to action $a$. This is described in detail in Section \ref{sec:app-box2} and illustrated in Figure \ref{fig:dt-as-policy}. 
(Learning a DT directly on a complex environment is often too time-consuming or difficult to be feasible. XAI-N enables utilizing an optimal initial learning method whilst taking advantage of tree structure after imitation.)
% We do not learn the DT directly since doing so on a complex environment can be difficult or time-consuming if not completely infeasible, and often results in policies with inferior performance compared to neural net policies. Learning a tree from an expert mitigates this because we can choose whatever initial learning method learns the most optimal policy, and gain benefits of the tree structure (after imitation) while mitigating the downsides. % PHD (compare to prev line, integrate)
Finally, the third stage is the modification stage where the policy is augmented. These augmentations, discussed in Section \ref{sec:app-box3}, can potentially improve the policy performance on several different navigation metrics, in some cases
beyond
that achieved by the expert policy. %We use the modification stage to improve several different navigation metrics. % By the end of the process, the policy has potentially combined the power of autonomous learning and human domain knowledge, and remains in an explainable, decision tree format at the end as well. PHD

\section{Approach: XAI-N Learning Method}\label{sec:app}

\subsection{Initial Learning Methods}\label{sec:app-box1} % (0.5 column)

The first stage can use any method of policy generation. For example, it could be created via any robot motion planning algorithm (eg. Sampling Based or Optimization-based algorithm)~\cite{lavalle2006planning}
or using reinforcement learning~\cite{sutton2018reinforcement}.
The important aspect of the first stage is that it can encompass any existing method that results in a robot policy.  We define an \textbf{expert policy} $\pi^* : (s \rightarrow a)$ as a function or object that maps from a state $s_i \in \mathcal{S}$ to an action $a_i \in \mathcal{A}, \forall s_i \in \mathcal{S}$ where $\mathcal{S}$ is the set of all possible states and $\mathcal{A}$ is a set of possible actions. Thus at every timestep $t$, the robot can observe state $s_t$, query $\pi^*$ to determine action $a_t$ to take, take that action, receive a new observation $s_{t+1}$, and repeat.  %The policy can have other attributes as well.

To generate our expert policy, we use Deep Reinforcement Learning, specifically Proximal Policy Optimization~\cite{PPO} in a Curriculum Learning (CL) pipeline.~\cite{bengio2009curriculum} We follow the procedure used in~\cite{fan2018crowdmove}.  
%Anyone could use a different policy generation method, and still replicate the remainder of our pipeline. %PHD

\subsection{Extraction to Decision Tree}\label{sec:app-box2} % (0.5 column)

\begin{figure}
    \centering
    \includegraphics[width=1\columnwidth]{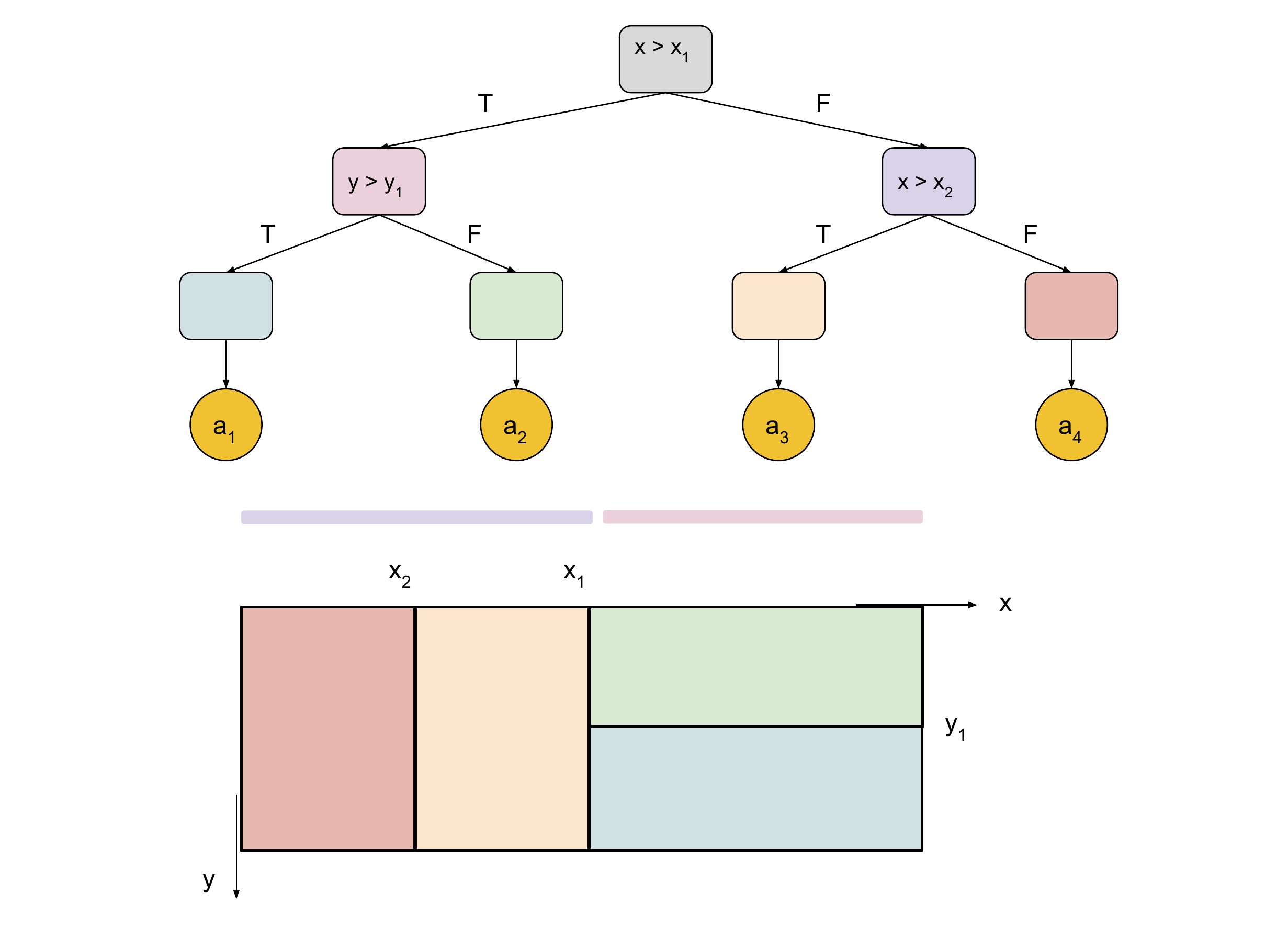}
    \caption{This figure demonstrates how a decision tree can be used as a policy for robot navigation. The rounded rectangles are branch and leaf nodes corresponding to abstract states. Each abstract state is a subset of the  robot's state space, and any given state $s$ falls inside the bounds of \textit{exactly} one leaf node abstract state (as well as the abstract states of all parent nodes to that leaf node). The root node has abstract space equivalent to the entire state space. Yellow circles are actions ($a_i$) (classes). In this example, a two-dimensional state-space is shown below the tree as a rectangle. The space is divided up by the tree. The colors demonstrate how the tree subdivides the state space into abstract states. A single $s_j$ would correspond to a single point within the bounds of the rectangle. Whichever leaf node / smallest divisible rectangle this state falls into will guide which action class the tree policy outputs for that state. This example uses arbitrarily chosen points of $x_1, x_2, y_1$ on two dimensions $x$ and $y$. Features could include meaning such as distance from or direction to a goal location, or the presence of obstacles, and the robot would take different actions in each case. Our actual environmental setup involves a policy with 213 state dimensions and 6 action classes (or 10 action classes after the oscillation-fixing procedure in Section \ref{sec:app-box3-oscillation}).  }
    \label{fig:dt-as-policy}
\end{figure}

In the second stage, the expert policy is transformed from its current format into a decision tree format, called the \textbf{extracted policy} $\pi^\dag$.  %If the expert policy is already a decision trees, $\pi^\dag = \pi^*$ directly, although such methods are rare in the literature.  More typically %(since non-decision tree methods are typically more performant than decision tree methods), 
We chose a decision tree as opposed to a regression algorithm because we want this stage to be more interpretable and modifiable.  % CR-rem
The conversion process $F^\dag$ converts $\pi^\dag = F^\dag(\pi^*)$.  Like $\pi^*$, $\pi^\dag$ maps from states to actions, but whereas the expert policy can have any internal structure (neural net, ensemble method, mixture of trees, planning algorithm, arbitrary code, etc) so long as it performs the $s \rightarrow a$ mapping, we constrain the decision tree to use a particular format, shown in Figure \ref{fig:dt-as-policy}.
The features of the decision tree correspond element-by-element to the features of the \textit{state space}, which is the term to describe $S$, or the space of all possible states.  A single state $s$ can be represented by an array of numbers. If the robot's raw observation is in a different format, such as a camera image, this input can be flattened or preprocessed into such an array. $S$ can be described by two arrays each equal in length to an $s$-array, and describing upper and lower bounds on the total state space.  A \textit{state subspace} (or an abstract state) is a subset of the state space, and can be similarly described by upper and lower bounds that demarcate a smaller space inside $S$. The elements of $s$ represent features, and these features are the features of the tree.  Each branching node of the tree thus splits on one feature of the state subspace, splitting it into two further subspaces, as shown.  Each leaf node's class label corresponds to an action (this can be a discrete action or an action probability distribution).

There are a number of ways to perform policy extraction (or policy distillation).  We use the VIPER family of methods because they result in a single tree policy and are applicable regardless of the internal structure of the expert policy~\cite{VIPER}.
% work regardless of the internal structure of the expert policy.~\cite{VIPER} 
In VIPER and its extensions, the expert policy $\pi^*$ is executed in the environment $E$. Each timestep, a state $s_t$ is observed and an action $a_t$ is chosen by $\pi^*$. We associate $(s_t, a_t)$ together as a ``state-action pair.'' The environment $E$ after receiving $a_t$ provides updated state $s_{t+1}$, and the cycle repeats until the goal is reached. This is called an ``episode,'' and multiple episodes are run, producing trajectories, or sequences of state-action pairs $\{(s_i, a_i), (s_{i+1}, a_{i+1}), ... \}$. These trajectories can  be combined into a dataset of state-action pairs.  Multiple datasets of state-action pairs are sampled from the total pairs generated. %This is a crucial step, which our implementation improves upon. Since our RL learner uses CL, we used a modified VIPER which samples trajectories from multiple stages instead of a single one. %CAMERA
These datasets are used in a supervised learning manner to learn a decision tree policy $\hat \pi _i$ using the CART method~\cite{lewis2000introduction}, with the state forming the features and the actions forming the labels. 
The resulting DT is a binary tree, where branch nodes test a condition regarding the feature space (which is the state space), and leaf nodes represent discrete action classes.
A diagram of this is shown in Figure \ref{fig:dt-as-policy}.  Whichever policy performs the best (as determined by which policy achieves the maximum average reward on a series of trials) is regarded as the best decision tree policy $\pi^\dag$. Reward is a property of environment $E$ and is constructed as a scalar that serves as a combined measure of the degree to which a robot is achieving certain navigation metrics. We use an $E$ where reward increases for reaching the goal or following a target, and doing so smoothly and quickly (described more in Section \ref{sec:eval-env}), such that $\pi^\dag$ most closely achieves the levels achieved by $\pi^*$ on these navigation metrics.
% Although this type of policy extraction has been been used with RL expert policies in the past, it has never been used on so complex an environment as ours. 

\subsection{Modification Methods}\label{sec:app-box3} % (4 columns)

In the third stage, we introduce modifications to improve the robot's ability to reach the goal without colliding with obstacles or freezing, to increase overall trajectory smoothness by reducing oscillation, and to avoid obstructing a human. Modifications targeting other navigation metrics could also be developed using a similar approach to what we have developed here.
% Given the decision tree structure, there is a large space of possible approaches to modification that are not be possible with other formats.  %PHD
% In particular, we consider that %PHD
A neural net format policy would not be able to be modified in the manner described in the following sections, hence the appeal of the DT.
% without first undergoing Extraction, as occurs in our method. PHD

% // BEGIN PHD
% A decision tree is interpretable and traceable. % Understanding the effects of a potential change do not involve tracing through mysterious weights and latent spaces, but simple observing the transparent structure of the tree.   PHD
% Algorithms can be created to modify a tree
% % , and in a development scenario, one can imagine engineers able to hand-craft desired modifications to a  %PHD
% policy without having to retrain it. %(This would be particularly useful for downstream users who received a policy which they have no ability to retrain.)  PHD
% In the following sections we describe a few modifications we created to demonstrate this.
% // END PHD

\subsubsection{Fix Freezing}\label{sec:app-box3-freezing}

One of the standard issues with navigation learning methods is ``freezing.'' % TODO-REF
The robot chooses to remain immobile in the face of certain obstacles.  Naturally, freezing helps prevent crashing, but %its also not desired behavior because 
the robot is also no longer moving towards the goal.  In particular, when the given obstacles are static, it is a failure mode from which it cannot escape.

We present a method to identify nodes in the tree that could be contributing to the freezing issue, and then modify those nodes to mitigate the danger of such an error occurring. The procedure for identifying nodes is shown in Algorithm \ref{alg:detect_freezing}

\RestyleAlgo{ruled}
\LinesNumbered
\begin{algorithm}%[H]
\textbf{Detect Freezing Nodes}( $\pi^\dag,  a_F, m_A$ ):\\
$\mathcal{N} \gets \varnothing$\;
\For {node $n \in \pi^\dag$}{
    \If {n\text{ is a leaf node}}{
        $m_C \gets$ the number of cells in the occupancy grid in which movement occurs\;
        \If {$m_C < m_A$ and $n[\text{action}] = a_F$} {
            Add $n$ to $\mathcal{N}$\;
        } 
    }
}
Return $\mathcal{N}$
 \caption{ Detect Freezing \label{alg:detect_freezing}}
\end{algorithm}

where $\pi^\dag$ is the tree policy, $a_F$ represents an action or grouping of actions corresponding to the ``stop'' action, and $m_A$ is a tunable integer parameter indicating ``in how many cells in the occupancy grid should movement be allowed while still declaring the obstacles stationary.'' (See section \ref{sec:eval-env} to explain the occupancy grid.) The algorithm checks each leaf node of the tree. If the obstacles detected are stationary within some tolerance indicated by $m_A$ and if the node's action is the Stop action, then the node is added to the list of problematic potential-freezing nodes. The $m_A$ parameter is included because in some situations we may not want to be completely strict about everything being perfectly still.
Setting $m_A = 0$ requires perfect stillness to consider a node a freezing possibility and setting $m_A$ to the maximum means the algorithm will return all nodes with the stop action $a_F$ regardless of obstacle position and movement. If a node's subspace dimensions encompass both moving and non-moving situations, the condition will be true for the purposes of this algorithm in the case that the bounds of those dimensions are unchanged for all timesteps (since even though movement could occur sometimes, the case where an obstacle is still is also included in this subspace).
The algorithm intended to alleviate this issue is found in Algorithm \ref{alg:fix_freezing},
where $a_R$ and $a_L$ are actions corresponding to pure right and left rotation (no linear velocity) respectively. This safely allows the robot to find an observed state where it can extract itself from stasis.
\RestyleAlgo{ruled}
\LinesNumbered
\begin{algorithm}%[H]
\textbf{Modify Freezing Nodes}( $\pi^\dag, \mathcal{N}, a_R, a_L$ ):\\
\For {node $n \in \mathcal{N}$}{
    \eIf{majority of obstacles are on the right}{
        $n[\text{action}] \gets a_L$\;
    }{
        $n[\text{action}] \gets a_R$\;
    }
}
Return the updated $\pi^\dag$
 \caption{ Alleviate Freezing \label{alg:fix_freezing}}
\end{algorithm}

% While there could theoretically still be freezing that occurs, in the form of rotation if trapped, this is still preferable to crashing. %  check grammar

\subsubsection{Fix Oscillation}\label{sec:app-box3-oscillation}

Another observed issue with some of the expert policies was oscillation. When seeking to circumvent certain obstacles, the robot would alternate between turning too far away from and towards the obstacle, resulting in aesthetically displeasing and inefficient behavior.  We developed a fix that involves running the policy in simulation and observing it to identify parts of the tree policy that contribute to the oscillation, and modifying the tree by adding nodes or modifying existing nodes to involve new actions with lower linear and angular velocities.  Detecting problematic nodes is done using Algorithm \ref{alg:detect_oscillation},
\RestyleAlgo{ruled}
\LinesNumbered
\begin{algorithm}%[H]
\textbf{Detect Oscillation Nodes}( $\pi^\dag, E, \mathcal{F}_O, n_e, L$ ):\\
$\mathds{H}^L \gets $ initialize an empty queue\;
$\mathcal{N} \gets \varnothing$\;
$\mathcal{O}_C(i) \gets \varnothing \quad \forall \quad i \in \{$ ids of nodes in $\pi^\dag\}$\;
$\mathcal{O}_X(i) \gets \varnothing \quad \forall \quad i \in \{$ ids of nodes in $\pi^\dag\}$\;
\For{$n_e$ episodes}{
    Reset environment $E$\;
    \While{E does not indicate episode done}{
        $s \gets$ get current state from $E$\;
        $a \gets \pi^\dag(s)$\;
        Append $(s, a)$ to $\mathds{H}^L$, removing the oldest if the length of the queue is $> L$\;
        \eIf{$\mathcal{F}_O(\mathds{H}^L)$}{
            \{$n_i, n_{i+1}, ..., n_{i+L}\} \gets $ leaf nodes in $\pi^\dag$ corresponding to each $s \in \mathds{H}^L$\;
            Add $\{n_i, n_{i+1}, ..., n_{i+L}\}$ to $\mathcal{N}$\;
            Add all $s \in \mathds{H}^L$ to $\mathcal{O}_C(d_i), \mathcal{O}_C(d_{i+1}), ..., \mathcal{O}_C(d_{i+L})$, where $d_d$ is the corresponding id of each node in $\{n_i, n_{i+1}, ..., n_{i+L}\}$\;
        }{
            $n_i \gets \text{ leaf node in $\pi^\dag$ corresponding to } s$\;
            Add $s$ to $\mathcal{O}_X(d)$, where $d$ is the id of $n_i$\;
        }
        Execute action $a$ in environment $E$\;
    }
}
Return $\mathcal{N}, \mathcal{O}_C(i), \mathcal{O}_X(i)$\;
 \caption{ Detect Oscillation \label{alg:detect_oscillation}}
\end{algorithm}
where $E$ is an environment, $\mathcal{F}_O$ is a function that takes in a history of state-action pairs and outputs a boolean indicating whether oscillation has occurred or not, $L$ is the length of that history, and $n_e$ is the total number of episodes to observe.
The modification procedure to correct this error is shown in Algorithm \ref{alg:fix_oscillation}. Nodes with subspaces that correspond to instances of oscillation are split, with the child leaf node corresponding to that subspace assigned a lower magnitude velocity action, and the sibling leaf node assigned the action of the original node.

\RestyleAlgo{ruled}
\LinesNumbered
\begin{algorithm}%[H]
\textbf{Alleviate Oscillation Nodes}( $\pi^\dag, \mathcal{N}, \mathcal{O}_C, \mathcal{O}_X, z$ ):\\
// Note that all $n \in \mathcal{N}$ are in $\pi^\dag$\\
\For{$n_i \in \mathcal{N}$}{
    \eIf{$\mathcal{O}_X(i) = \varnothing \text { or } z$}{
        // All states visited on this node are oscillation\\
        $n_i$[action] $\gets$ action with linear and angular velocity of reduced magnitude\;
    }{
        $X \gets$ new leaf node with action $n_i$[action]\;
        $C \gets$ new leaf node with action with linear and angular velocity of reduced magnitude compared to $n_i$[action]\;
        $n_i$ is turned into a branch node with $X$ and $C$ as children, splitting on the best split that best separates the states in $\mathcal{O}_C(i)$ to node $C$ and states in $\mathcal{O}_S(i)$ to node $X$\;
    }
}
Return updated $\pi^\dag$\;
 \caption{ Alleviate Oscillation \label{alg:fix_oscillation}}
\end{algorithm}
In Algorithm \ref{alg:fix_oscillation}, each $\mathcal{O}_C(i) \in \mathcal{O}_C$ is a set of states in state subspace of node $i$ where oscillation occurs, and each $\mathcal{O}_X(i) \in \mathcal{O}_X$ is a set of states in state subspace of node $i$ where oscillation does not occur, and $z$ is a boolean.

\subsubsection{Fix Blocking/Obstruction}\label{sec:app-box3-blocking}

In the warehouse environment, where the robot locates and follows a human, we found that the robot sometimes would place itself in the human's path, blocking the human.  This is inefficient, and would be annoying or dangerous in real life.
% We developed an algorithm to detect when the robot might be in this situation and to resolve it by extracting itself from it.  %PHD
Find the algorithm used to detect potential nodes contributing to this situation in Algorithm \ref{alg:detect_blocking}, 
\RestyleAlgo{ruled}
\LinesNumbered
\begin{algorithm}%[H]
\textbf{Detect Blocking Nodes}( $\pi^\dag,  \mathbf{A}_D, m_A$ ):\\
$\mathcal{N} \gets \varnothing$\;
\For {node $n \in \pi^\dag$}{
    \If {n\text{ is a leaf node}}{
        $m_C \gets$ the number of cells in the occupancy grid in which movement occurs\;
        \If {$m_C < m_A$ and $n[\text{action}] \notin \mathbf{A}_D$} {
            Add $n$ to $\mathcal{N}$\;
        } 
    }
}
Return $\mathcal{N}$
 \caption{ Detect Blocking \label{alg:detect_blocking}}
\end{algorithm}
where $\pi^\dag$ is the tree policy, $a_F$ represents an action or grouping of actions corresponding to the ``stop'' action, $m_A$ is a tunable parameter and $\mathbf{A}_D$ is the set of actions that imply no blocking is occurring (ie all the movement actions with a forward component.
% since if the robot is moving, it is not blocked or blocking). %CAMERA
The algorithm intended to alleviate this issue is found in Algorithm \ref{alg:fix_blocking},
\RestyleAlgo{ruled}
\LinesNumbered
\begin{algorithm}%[H]
\textbf{Modify Blocking Nodes}( $\pi^\dag, \mathcal{N}, a_L, a_R, a_D, \mathbf{C}_F, n_r$ ):\\
\For {node $n \in \mathcal{N}$}{
    $\text{clear\_in\_front } \gets$ a boolean true if the columns in $\mathbf{C}_F$ are clear for a distance of $n_r$ rows, starting from the nearest row\;
    \eIf{$\text{clear\_in\_front}$}{
        $n[\text{action}] \gets a_D$\;
    }{
        \eIf{majority of obstacles are on the right}{
            $n[\text{action}] \gets a_L$\;
        }{
            $n[\text{action}] \gets a_R$\;
        }
    }
}
Return the updated $\pi^\dag$
 \caption{ Alleviate Blocking \label{alg:fix_blocking}}
\end{algorithm}
where $a_R$ and $a_L$ are the right and left rotation actions, $a_D$ is the ``forward'' action, $\mathbf{C}_F$ are indices of the columns of the polar grid directly in front of the robot (encompassing a traversable expanse, such that the robot could proceed forward into that region without collision), and $n_r$ is how far ahead to look in number-of-rows when checking whether those columns are occupied.

In this manner, the robot's policy is changed so that in situations where it might be stuck, it seeks open space and moves there, presumably away from a near obstacle which may or may not be a human.

\section{Evaluation}\label{sec:eval}
\subsection{Environments}\label{sec:eval-env} % (0.5 column)

We demonstrate our improvements on two environments.
\subsubsection{Mobile Robot Navigation}\label{sec:results-env}
The robot starts in a random location and must navigate around obstacles to a random goal location. 
Obstacles can be static or dynamic. % with relatively slow speed, which could be captured by robots' sensors.
We desire that the policy should perform well in terms of avoiding the pedestrians and obstacles.
We created a simulation of this environment and also test in a real-world setup, in both cases with a Clearpath Jackal. We formulate the environment as an AI Gym~\cite{brockman2016openai}, a common RL interface for environments, and release it as open-source code for others to use as well\cite{crowdenv_github}.
% \footnote{\url{https://github.com/AMR-/JackalCrowdEnv}}

\subsubsection{Game Character Locomotion and Animation}

In this environment, the agent is a character which spawns in a complex multi-room environment with obstacles and other characters with which to interact. %Our goal is to generate smooth and natural looking animation as the agent completes the task.
There are three stages in this game: i) learning to exit the room, ii) learning to exit the room and finding another certain autonomous character, iii) following this other character as they move.

\subsubsection{Sensors and State Space}

The sensor setup in both cases involves lidar and a pozyx system (an ultra-wideband based localization system)~\cite{mimoune2019evaluation}.
The state space contains information about the goal location (relative to the robot) in polar coordinates, the previous robot action, and the physical surroundings of the robot as sensed by the lidar.  The lidar we use scans $512$ ranges from $-\frac{2}{3}\pi$ to $\frac{2}{3}\pi$ radians (with $0$ radians corresponding to straight ahead).  We transform this into a radial occupancy grid. % Although our approach could be used with an arbitrary occupancy grid, PHD
In our implementation of this benchmark we use a grid with 10 evenly spaced columns, and rows start 10 cm from the center of the robot, with distances of the 7 rows as (listed in order from nearest to farthest from the robot) $0.2$ m, $0.2$ m, $0.2$ m, $0.3$ m, $1$ m, $1$ m, $1$ m. 
The state space contains the occupancy grid information from the current time step and previous two timesteps. There are thus 210 features describing obstacle position and movement, 2 features indicating relative goal position, and 1 feature indicating the previous action chosen by the agent (for a total of 213 features).
The action space is a discrete action space: 
1) Forward and Left, 2) Rotate Left, 3) Straight Forward, 4) No movement, 5) Forward and Right, 6) Rotate Right.  (We also implemented an expanded action space that contains four additional actions that correspond to actions 1, 2, 5, and 6 but with smaller magnitude velocities.) 
\begin{table}[]
    % \centering
    \resizebox{0.5\textwidth}{!}{
    \begin{tabular}{|c|c|c|c|c|c|c|c|c|c|}
         \hline
        \makecell{XAI-N\\Stage} & \makecell{Policy\\Type} & \makecell{Avg \\ Reward
        \\per\\timestep} & \makecell{\% \\ crash} & \makecell{\% \\freeze} & \makecell{Oscill-\\ation \%} & \makecell{Avg\\Osc. \\length} & \makecell{Path\\Length(m)} \\
        \hline
        % - & RV0 & - & 0 & 1 & 0 & 0 & 8.66 \\
        %  \hdashline
        1 & Expert (PPO) & 0.226 & 0\% & 0\% & 100.\% & 8.07 & 9.94 \\
        \hdashline
         % saved_clf_2020-10-09_23-23-44_3.pkl
        2 & M-VIPER\textsuperscript & -0.276 & 67\% & 0\% & 95\% & 1.73 & 8.18 \\
        % \hdashline
        \hdashline
        % saved_tree_2020-10-09_23-23-44_3_oscmod_20201027_2321.pkl
        3 & \makecell{M-VIPER +\\Oscillation Fix (XAI-N)} & \textbf{0.241} & 4\% & 0\% & \textbf{6}\% & \textbf{1.33} & 8.29 \\
      %  \hdashline
        % saved_tree_2020-10-09_23-23-44_3_oscmod_20201027_2321.pkl
        %4 & \makecell{DWA} & NAN & 10\% & 0\% & \textbf{0}\% & \textbf{0} & 11.20 \\
        \hline
    \end{tabular}
    }
    \caption{
        Policy at different stages of our XAI-N algorithm.  At the intermediate stage 2  (Figure \ref{fig:paradigm}), average reward per timestep decreases due to crashing, this is resolved by stage 3, where crashing decreases, average reward per timestep is higher than preceding stages including stage 1 expert policy, and the 100\% oscillation from stage 1 is reduced to 6\%. Overall, we design an improved learning-based navigation algorithm.
        %We also compare and demonstrate the performance improvement over well-known dynamic-window algorithm (DWA)~\cite{dwa}, which has a higher crash rate.
    }
    % \caption{%Here are some results, from scenario 10.\\
    % \textsuperscript{A} Best M-VIPER. \textsuperscript{B} The M-VIPER that is base for best oscillation fix.  \textsuperscript{C} The M-VIPER that is base for best freezing fix
    % }
    \label{tbl:results}
    \vspace*{-6mm}
\end{table}

We design the reward function with three major parts, as follows:
\begin{equation}
    \textit{r}_i^t = (r_g)^t_i \,+ (r_c)^t_i \,+ (r_{o})^t_i. 
\end{equation}
where $(r_g)_i^t$ rewards movement towards and reaching the goal or person, $(r_c)_i^t$ penalizes collisions with or proximity to obstacles, $(r_o)_i^t$ penalizes oscillations and rewards smoothness. %Details on each of these subcomponents can be found in documentation in the repository. % TODO DOCUMENTATION

We used Gazebo 9.0 simulator with ROS Melodic on Ubuntu 18.04 to create multiple scenarios with different types and layouts of the obstacles. 

\begin{table}[b]
    \centering
    \begin{tabular}{|c|c|c|c|c|c|c|c|c|c|c|c|c|c|}
         \hline
        \makecell{XAI-N Stage} & \makecell{Policy Type} & \makecell{\% freezing} \\
        \hline
        % TODO-Jing which tree files?
         2& \makecell{M-VIPER} & 28\%\\
         \hdashline
         3& \makecell{M- VIPER + Freezing Fix} & 0\% \\
        \hline
    \end{tabular}
    \caption{Our method detected freezing after imitation learning in a stage 2 M-VIPER.  We ran the freezing fix to correct this issue. Our resulting learning-algorithm exhibits no freezing issues.
    }
    \label{tbl:freezing_result}
\end{table}

\begin{table}[b]
    \centering
    \begin{tabular}{|c|c|c|c|c|c|c|c|c|c|c|c|c|c|}
         \hline
        \makecell{XAI-N Stage} & \makecell{Policy Type} & \makecell{Avg \% blocking} \\
        \hline
         % game msv saved_clf_2021-02-04_22-49-43_1.pkl
         2& \makecell{M-VIPER} & 0.7 \\
         \hdashline
         % saved_tree_2021-02-04_22-49-43_1_noblock3_20210305_0126.pkl
         3& \makecell{M- VIPER + Blocking Fix} & 0.0 \\
        \hline
    \end{tabular}
    \caption{Our method detected potential blocking behavior after imitation learning in the warehouse environment.  We ran the blocking fix to correct this issue and our learning algorithm exhibits no blocking behavior.
    }
    \label{tbl:blocking_result}
\end{table}

\subsection{Results}\label{sec:eval-results} % (2 columns)

Find our detailed results in Table \ref{tbl:results}, and you can also view the accompanying video for a live demonstration.
CrowdEnv scenario 10 was used to test and produce the data.
%``Env Scenario'' indicates which scenario (obstacle configuration) of the CrowdEnv environment was used to test. 
Scenario 10 was \textbf{not} a configuration of obstacles that any of the policies saw during their training. ``Path Length'' is an average of total path lengths, counting only those runs where the robot successfully reached the goal.
The expert policy $\pi^*$ (labeled ``PPO'' in reference to the DRL training method used) demonstrates an average of over 8 meters of oscillation per run, and oscillated during every run. The policy after the conversion to decision tree $\pi^\dag$ is noted as Modified VIPER (M-VIPER).  Generally the fixed decision tree inherits the optimality regarding path length of training based algorithm and also improves the performance of navigation regarding the specific issues of decreasing crash rate, freeze rate and oscillations where they occur. %Comparing with Dynamic Window Approach~\cite{dwa}, to achieve the similar Path length we have less crashes as Table \ref{tbl:results}.
%Also shown for comparison are the performance of a policy trained using RV0. % TO DO-REF
% RV0 underperforms the expert policy and and final fixed policies in terms of average reward per timestep, and while it has a shorter path than the PPO policy, it has a longer path than the M-VIPER and final fixed policies. % CAMERA or PHD, prev two lines
% Our results for the various fixes are shown in separate tables because these modifications are not generic improvements to be run in sequence but rather targeted interventions.

\begin{figure}
    \centering
    \includegraphics[width=0.29\columnwidth]{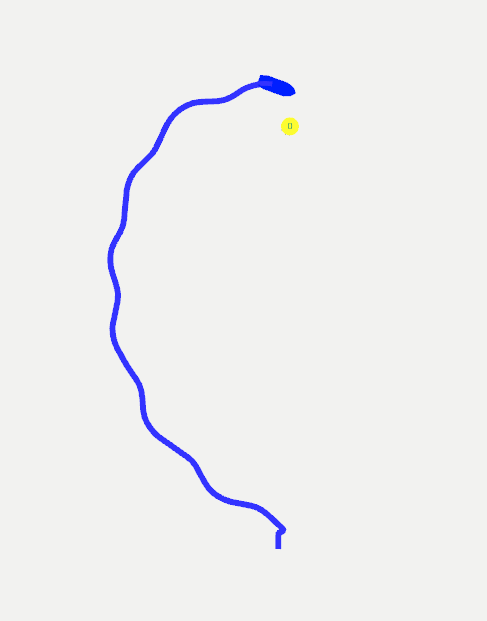} \includegraphics[width=0.29\columnwidth]{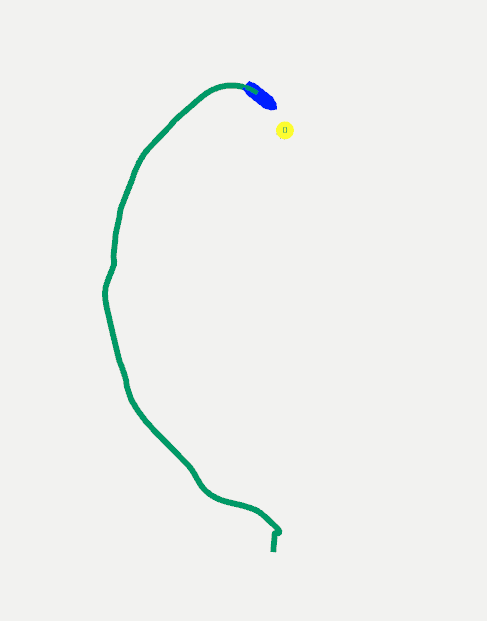} \includegraphics[width=0.37\columnwidth]{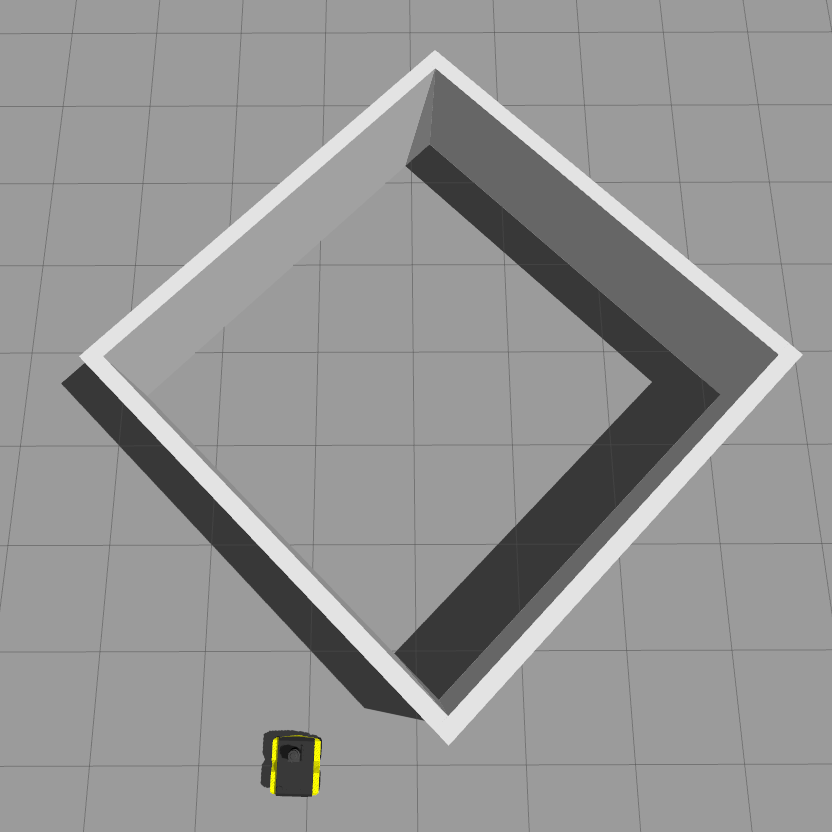}
    \caption{At right, an obstacle the robot must circumnavigate. At left, the robot's path before the oscillation fix is applied (i.e, DRL expert policy). At center, the robot's path after oscillation fix is applied using policy extraction and DTs.}
    \label{fig:oscillation}
    \vspace*{-4mm}
\end{figure}

{\bf Reduced Oscillations:} To demonstrate the oscillation fix, we chose one of the extracted $\pi^\dag$ that still had a significant amount of oscillation after extraction.  Labeled as ``M-VIPER'', we see it has an oscillation 95\% of the time, and an oscillation length of 1.73.   Firstly, something interesting to note is that the extraction process itself reduced the length of oscillation significantly. 
%We theorize that this is due to the abstract states. 
At this intermediate stage, the average reward per timestep decreases since it crashes more than expert, due to imperfect imitation.
%TODO - note specific parameters of the VIPER trainings used
After performing our oscillation fix (with $n_e=20, L=5, z=$true)
, we obtain the policy shown in the ``M-VIPER + Oscillation Fix'' (i.e. after Stage 3 of our XAI-N approach), The oscillation fix procedure identified 11 nodes in the DT that might be contributing to oscillation, and applied the fix to them. The crashing issue is resolved. In the policy, oscillation occurs in only 6\% of runs, and has a oscillation length reduced to an average cumulative of $1.33$ m in those rare instances where it does occur. Find an illustration in Figure \ref{fig:oscillation}. This is a significant reduction in oscillation beyond that achieved by the standalone DRL method, despite the fact that the reward function for the DRL method included a parameter to reward smoothness (decreasing oscillation).  
%TODO consider adding back in the following
%This is probably an explanation for why the fixed policy achieves a higher average reward per timestep than the DRL policy (0.241 compared to 0.226). 
%Interestingly, achieving the smoother path apparently increased the average path length, from 8.18 to 8.29, although this is still shorter than what was achieved by the base PPO policy.

{\bf Eliminate Freezing:} We applied the freezing fix to a different M-VIPER policy as shown in Table \ref{tbl:freezing_result}. This policy would freeze 28\% of the time.
% not shown in the table.  This policy would freeze 92\% of the time.
The freezing fix identified 30 nodes that may have contributed to the error, and modified them accordingly (out of a total of 621 nodes, 311 of which are leaf nodes).
After applying the freezing fix, freezing was eliminated using our XAI-N approach. 
% (although in those cases where it no longer froze, it was still sometimes unable to reach the goal).
% --whether this is superior to immobility would be application dependent, and is still a useful option to have).

{\bf Eliminate Blocking:} We applied the blocking fix to a warehouse policy that exhibited blocking, as shown in Table \ref{tbl:blocking_result}. The blocking fix identified 381 nodes to potentially modify out of 1559 nodes total. %TODO check this
151 nodes were adjusted to move the robot forward, and the other 230 were given rotation actions to orient the robot in a manner where it could more safely move out of the human's way.  %After applying the blocking fix, blocking was eliminated using our XAI-N approach.

The blocking fix is an example of a case of trade-offs.  Although blocking was eliminated, it decreased the efficiency of the path (increasing average trajectory length from 9.45 to 17.3).  In the pure navigation environment, this would not be desirable.  The warehouse environment, however, simulates a human-robot interaction scenario, and in this situation one can imagine a preference for safety and comfortable robot interaction, in comparison to ``most-efficent'' paths that might nip at a humans' heels or obstruct the human's path.
This kind of domain-specific customization, based around a similar sensor scheme for a robot and similar learning procedures, demonstrates the usefulness of our paradigm.

\section{Conclusion and Future Work}\label{sec:eval-conclusion} % (0.5 column) 

We provide XAI-N, an improved learning algorithm for sensor-based robot navigation. Starting with training an expert policy (e.g trained by DRL), we extract a decision tree policy, the % explainable and 
interpretable properties of which we utilize to modify the tree. This allows for improving smoothness of path, mitigating the chance of obstructing a human, and reducing the problem of freezing. %One of the limitations of this method is that the abilities of the it is constrained by the  %One of the limitations of this method is that it requires human input. While in some ways this is a benefit (in that it allows for customized, human input), it would also be useful to develop
We are able to modify the policy to address these imperfections without retraining, combining the learning power of deep learning with the control of domain-specific algorithms. We demonstrated fixes across two environments, a robot navigation among pedestrians and obstacles, and a warehouse game with an agent following a person.

One limitation is that the maximum speed of the dynamic obstacles should not be more than the maximum speed of the robot itself.
Future work could address this, could include modification techniques for tackling additional navigation issues beyond freezing, oscillation, and blocking, or could combine XAI-N with other motion planning methods\cite{manocha1992algebraic}.

% \section*{Acknowledgement}

% conclusion  - partic useful in real world development, where the purity of a particular learning method not as important as solving the overall problem by whatever means necessary %PHD

\bibliographystyle{./bibtex/IEEEtran}
\bibliography{./bibtex/bibtex}

\end{document}